\documentclass{article}
\usepackage{spconf,amsmath,graphicx}
\usepackage{booktabs}

\usepackage{multirow}
\usepackage{enumitem}
\setlist{nosep, leftmargin=14pt}

\usepackage{mwe} 

\title{AID-DTI: Accelerating High-fidelity Diffusion Tensor Imaging with Detail-Preserving Model-based Deep Learning}
\name{%
\begin{tabular}{c}
    {Wenxin Fan$^{1,2}$ \qquad Jian Cheng$^{3,4}$ \qquad Cheng Li$^{1}$ \qquad Xinrui Ma$^{1,2,}$ \qquad Jing Yang$^{1,2,}$ } \\
    {Juan Zou$^{1}$ \qquad Ruoyou Wu$^{1,2,5}$ \qquad Qiegen Liu$^{6}$ \qquad Shanshan Wang$^{1,2,5}$}
  \end{tabular}
}

\address{$^{1}$Paul C. Lauterbur Research Center for Biomedical Imaging,\\ Shenzhen Institute of Advanced Technology, Chinese Academy of Sciences, Shenzhen, Guangdong, China \\
    $^{2}$University of Chinese Academy of Sciences, Beijing, China\\
    $^{3}$State Key Laboratory of Software Development Environment, Beihang University, Beijing, China. \\
    $^{4}$Key Laboratory of  Data Science and Intelligent Computing, \\Institute of International Innovation, Beihang University, Hangzhou, Zhejiang, China\\
    $^{5}$Peng Cheng Laboratory, Shenzhen, Guangdong, China
    $^{6}$Nan Chang University, Nan Chang, Jiang Xi, China\\
}

\begin{document}
%
\maketitle
\begin{abstract}
Deep learning has shown great potential in accelerating diffusion tensor imaging (DTI). Nevertheless, existing methods tend to suffer from Rician noise and detail loss in reconstructing the DTI-derived parametric maps especially when sparsely sampled q-space data are used. This paper proposes a novel method, AID-DTI (\textbf{A}ccelerating h\textbf{I}gh fi\textbf{D}elity \textbf{D}iffusion \textbf{T}ensor \textbf{I}maging), to facilitate fast and accurate DTI with only six measurements. AID-DTI is equipped with a newly designed Singular Value Decomposition (SVD)-based regularizer, which can effectively capture fine details while suppressing noise during network training. Experimental results on Human Connectome Project (HCP) data consistently demonstrate that the proposed method estimates DTI parameter maps with fine-grained details and outperforms three state-of-the-art methods both quantitatively and qualitatively.
\end{abstract}
\section{Introduction}
\label{sec:intro}

Diffusion magnetic resonance imaging (dMRI) is a prominent non-invasive neuroimaging technique for measuring tissue microstructure. Among various dMRI techniques, diffusion tensor imaging (DTI) \cite{basser1994mr} is the most widely used method to extract brain tissue properties and identify white matter tracts in vivo. The metrics from DTI have great specificity in mapping the microstructural changes caused by normal aging \cite{salat2005age}, neurodegeneration \cite{thompson2013effectiveness}, and psychiatric disorders \cite{zheng2014dti}, such as fractional anisotropy (FA), mean diffusivity (MD), and axial diffusivity (AD) \cite{curran2016quantitative}. 
To increase the accuracy of DTI-derived parametric maps, studies typically need more than the minimum of 6 diffusion weighting (DW) directions or acquire repeated observations of the same set of DW directions \cite{landman2007effects}. Moreover, diffusion MRI data often suffer from Rician noise, which further increases the demand for data to enable high-fidelity DTI metrics, which is time-consuming and vulnerable to artifacts. There is an urgent need to develop high-quality DTI metrics estimation from sparsely sampled q-space data.

Recently, deep learning has emerged as a powerful tool for accelerating DTI imaging. The pioneering work, q-space deep learning (q-DL) \cite{golkov2016q}, was introduced to directly map a subset of diffusion signals to Diffusion Kurtosis Imaging (DKI) parameters using a three-layer multilayer perceptron (MLP). Gibbons et al. \cite{gibbons2019simultaneous} used a 2D convolutional neural network (CNN) to estimate the Neurite Orientation Dispersion and Density Imaging (NODDI) and generalized fractional anisotropy maps. In addition to data-driven mapping approaches, there has been a growing interest in model-driven neural networks that leverage domain knowledge to enhance the network performance and interpretability. A notable example is the works proposed by Ye et al. \cite{ye2019deep, ye2020improved, zheng2023microstructure} which unfold the iterative optimization process for parameter mappings. Chen et al. used a subset q-space to estimate the parameters by explicitly considering the q-space geometric structure with a graph neural network (GNN) \cite{chen2020estimating, chen2022hybrid}.

Despite the progress made, the current methods still suffer from noise corruption or fine detail loss at highly accelerated imaging rate. In this study, we propose a novel model-based deep learning method, named \textbf{AID-DTI} (\textbf{A}ccelerating h\textbf{I}gh fi\textbf{D}elity \textbf{D}iffusion \textbf{T}ensor \textbf{I}maging). The main contributions of this work can be summarized as follows:
\begin{enumerate}
\item We propose a simple but effective model-based deep learning model, with a newly designed regularization term to facilitate high-fidelity DTI metrics derivation by explicitly constraining the alignment of primary singular values, effectively suppressing noise, and preserving important details.  
\item The data requirement of AID-DTI is simplified and minimized to six measurements along uniform diffusion-encoding directions to help accelerate DTI metrics estimation.
\item Our method outperforms three state-of-the-art methods both quantitatively and qualitatively.
\end{enumerate}

\section{Method}
\label{sec:format}
In this section, we present the details of AID-DTI, which investigates the SVD-based regularization to estimate accurate DTI metrics using only six measurements instead of the recommended 30 measurements to achieve reliable DTI metrics within the needed clinical accuracy \cite{jones2013white}. Additionally, our approach automates the adjustment of regularization parameters during the training process.
\subsection{Task Formulation}
Given the diffusion MRI data $X$ containing the full measurements in the q-space, the ground-truth scalar maps $Y_{GT}$ obtained from all the diffusion data, we aim at finding a mapping ${\mathcal{F}}$ s.t. $\mathcal{F}(\tilde{X})=Y\rightarrow Y_{GT}$ where the given sparse sampling data $\tilde{X}\subseteq X$, the overall optimization problem used to be defined as:$$\begin{aligned}\min_{\tilde{X}\subseteq X}L(\mathcal{F}(\tilde{X}),Y_{GT})+R(Y)\end{aligned}$$ where $L(\cdot)$ denotes the loss function, and $R(\cdot)$ is the regularization over the given predicted metrics.

\subsection{SVD-based Regularization}
\begin{figure}[htb]
  \centering
  \centerline{\includegraphics[width=9cm]{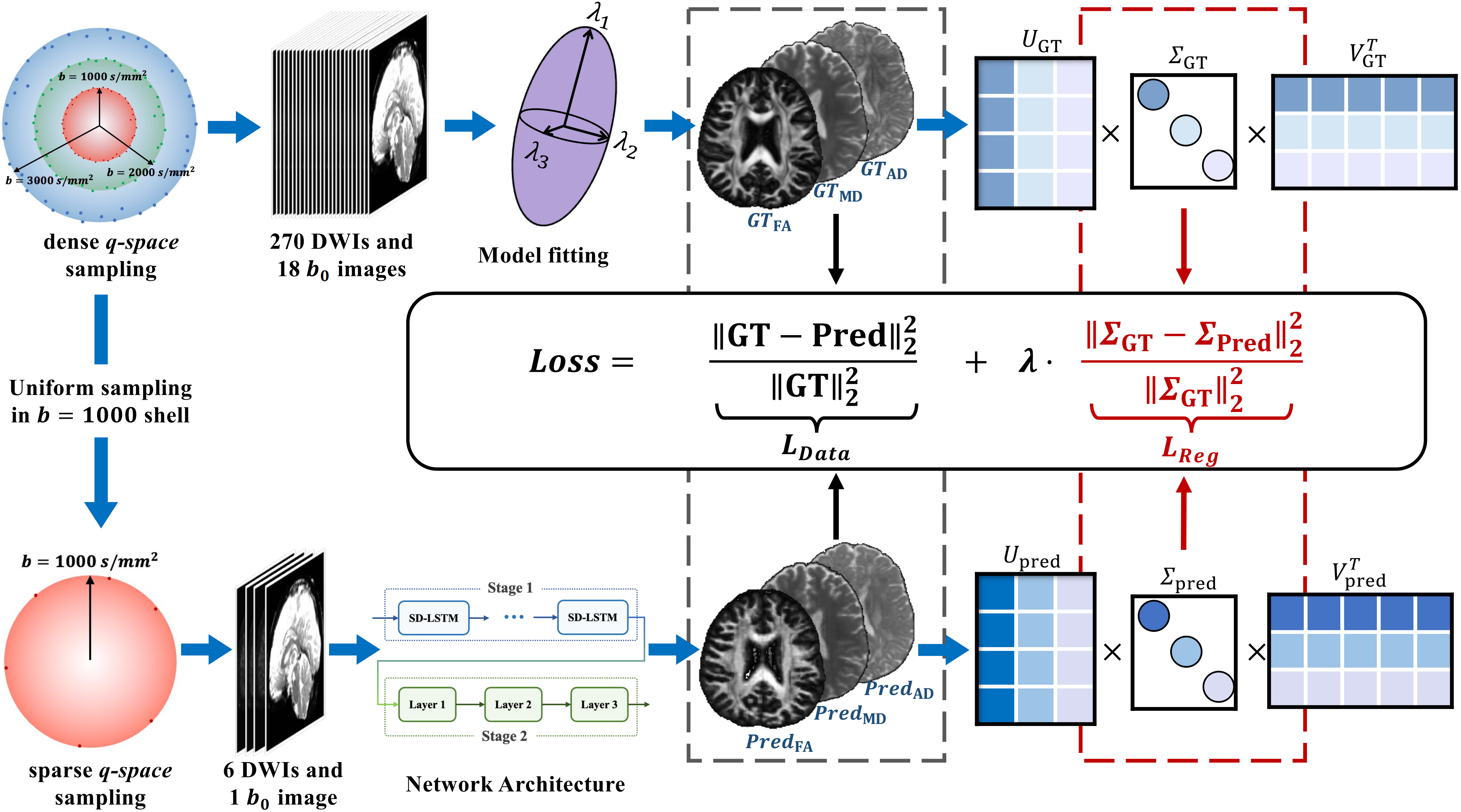}}
  \caption{Diffusion physics-informed AID-DTI pipeline. The input is a single $b=0$ image and six DWI volumes sampled along uniform diffusion-encoding directions. The output is the high-quality volume of DTI metrics estimation. }
\label{fig1}
\end{figure}

The overall framework of AID-DTI is shown in Figure \ref{fig1}. The inputs are: a single non-diffusion-weighted ($b=0$) image volume and six diffusion-weighted image (DWI) volumes sampled along uniform diffusion encoding directions (a total of seven input channels). Then the mapping between the sparse sampling and high-fidelity DTI metrics is directly learned. It is worth noting that, after obtaining the network’s output, instead of simply calculating $L_2$ loss to optimize the network, we performed SVD operation on both the prediction and the ground truth, ensuring consistency in the singular value matrix as well. Given the ability of SVD to extract the most dominant features, we propose the incorporation of an SVD-based regularization term to ensure consistency between the prediction and the ground truth on the singular value matrix as well. 

Based on the deep supervision structure, we design a deeply supervised SVD-based loss, which constrains the data consistency across the spatial and singular-value space. Our task-oriented loss function is defined as:$$\begin{aligned}
\text{\textit{Loss}}& =L_{Data}+\lambda\cdot L_{Reg}  \\
&=\frac{\|\mathrm{GT}-\mathrm{Pred}\|_2^2}{\|\text{ GT }\|_2^2}+\lambda\cdot\frac{\left\|\Sigma_{\mathrm{GT}}-\Sigma_{\mathrm{Pred}}\right\|_2^2}{\left\|\Sigma_{\mathrm{GT}}\right\|_2^2}
\end{aligned}$$ 
where $\Sigma_{\mathrm{GT}}$ and $\Sigma_{\mathrm{Pred}}$ represent the ground-truth metrics and their singular value diagonal matrix of Ground Truth metrics and predicted data respectively.

\subsection{Rationale of the Proposed Regularization}
\begin{figure}[htb]
  \centering
  \centerline{\includegraphics[width=9cm]{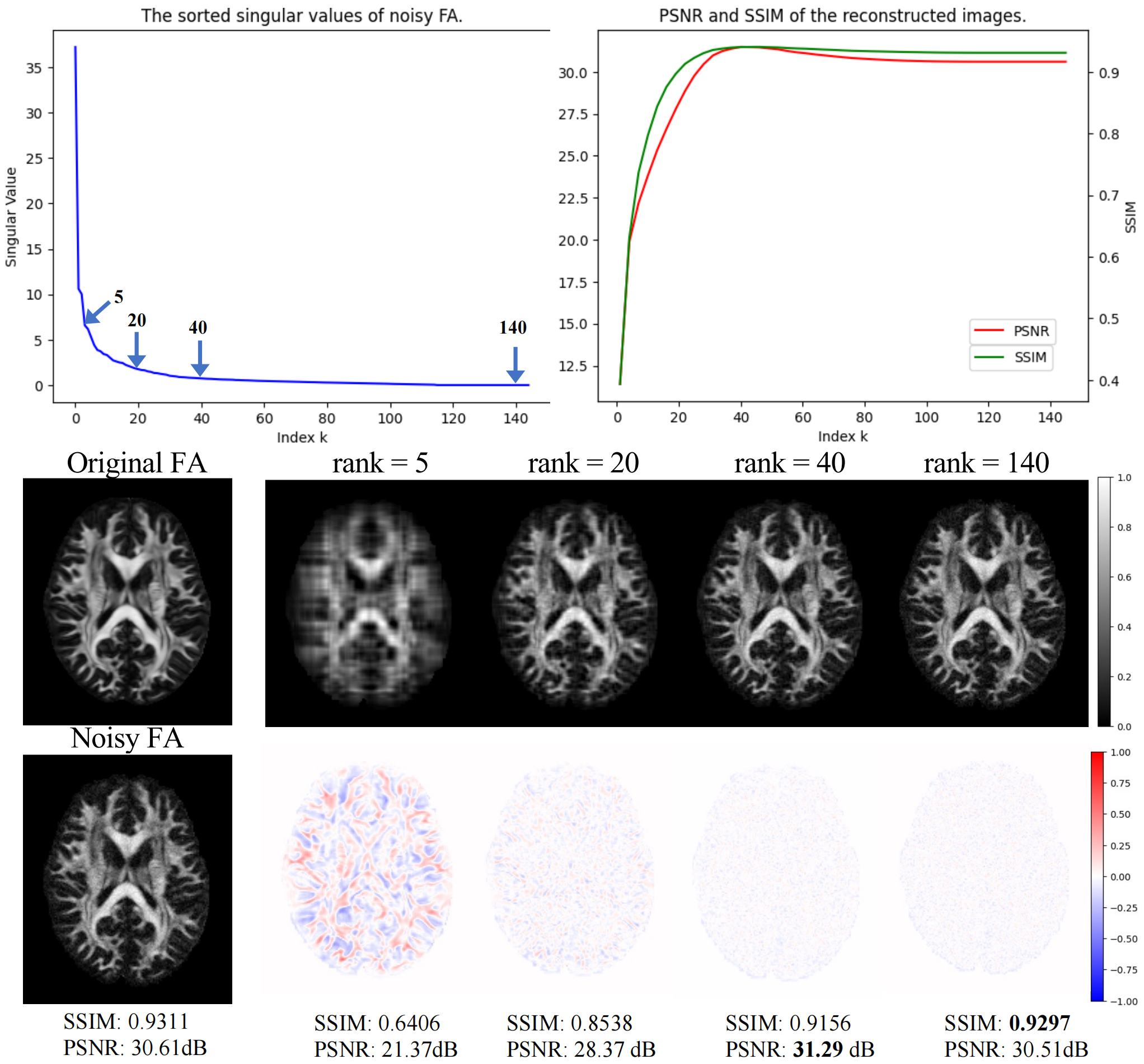}}
  \caption{The upper row displays the plot of singular values along with the trends of PSNR and SSIM, showcasing how they vary with the number of selected singular values. The lower row exhibits the original FA image, the noisy FA image, as well as four reconstructed images and corresponding error maps obtained from the noisy FA using SVD, where the top 5, 20, 40, and 140 singular values are employed, respectively.}
\label{fig2}
\end{figure}

According to the Eckart–Young theorem \cite{eckart1936approximation}, the dominant singular subspaces (i.e., the span of the singular vectors corresponding to the largest singular values) capture the majority of the informational content. Therefore, aligning the primary singular values can effectively preserve fine details while removing a certain degree of noise.
We have shown a toy example in Figure \ref{fig2}, which shows singular values decrease dramatically, with most values being close to zero. When Rician noise corrupts the original FA image, details may get tangled with the noise. The upper right of Figure \ref{fig2} demonstrates the trend of peak signal to noise ratio (PSNR) and structural similarity index (SSIM) of the reconstructed images with our proposed method, when the original data was corrupted by a Rician noise with $\sigma=0.04$, SNR=25, for example. From this plot, we can observe that as the number of singular values increases, the reconstructed images exhibit richer information, leading to a continuous development in both PSNR and SSIM. However, after reaching a certain critical point (around $rank=40$ in this case), while enriching the structural details, noise is also introduced, resulting in a decrease in PSNR while SSIM continues to increase. As the singular values approach zero, containing little meaningful information, both PSNR and SSIM tend to stabilize. Then, we remove the small singular values and use the top 5, 20, 40, and 140 singular values to reconstruct the image. The lower part of Figure \ref{fig2} displays the reconstructed images along with their corresponding error maps.

\subsection{Network Architecture}
Note that AID-DTI is a flexible framework and can utilize any network architecture suitable for the task. Here, the Microstructure Estimation with Sparse Coding using Separable Dictionary (MESC-SD), an unfolding network based on sparse LSTM units \cite{zhou2018sc2net} with two cascaded stages, is employed. The first stage computes the spatial-angular sparse representation of the diffusion signal while the second stage maps the sparse representation to tissue microstructure estimates. 

\section{Experiments and Results}
\label{sec:pagestyle}

\subsection{Dataset}

Pre-processed whole-brain diffusion MRI data from the publicly available Human Connectome Project (HCP) dataset were used for this study \cite{van2013wu}. Diffusion MRI data were acquired at 1.25mm isotropic resolution with four b-values (0, 1000, 2000, 3000 $s/mm^{2}$). For each non-zero b-value, 90 DWI volumes along uniformly distributed diffusion-encoding directions were acquired. To obtain the input data of AID-DTI, DWI volumes acquired along six uniform diffusion-encoding directions at $b=1000s/mm^{2}$ of each subject were selected. To obtain the ground-truth DTI metrics, diffusion tensor fitting was performed on all the diffusion data using ordinary linear squares fitting implemented in the DIPY\footnote{https://github.com/dipy} software package to derive the FA, MD, and AD \cite{curran2016quantitative}.

\subsection{Experiment Settings}
The training was performed on the data of 60 subjects and tested on another 17 subjects, using four Tesla V100 GPUs (NVIDIA, Santa Clara, CA) with 32GB memory. The neural network was implemented using the PyTorch library (codes will be available online upon acceptance of the paper).

\subsection{Results and Discussion}
\begin{figure}[htb]
  \centering
  \centerline{\includegraphics[width=9cm]{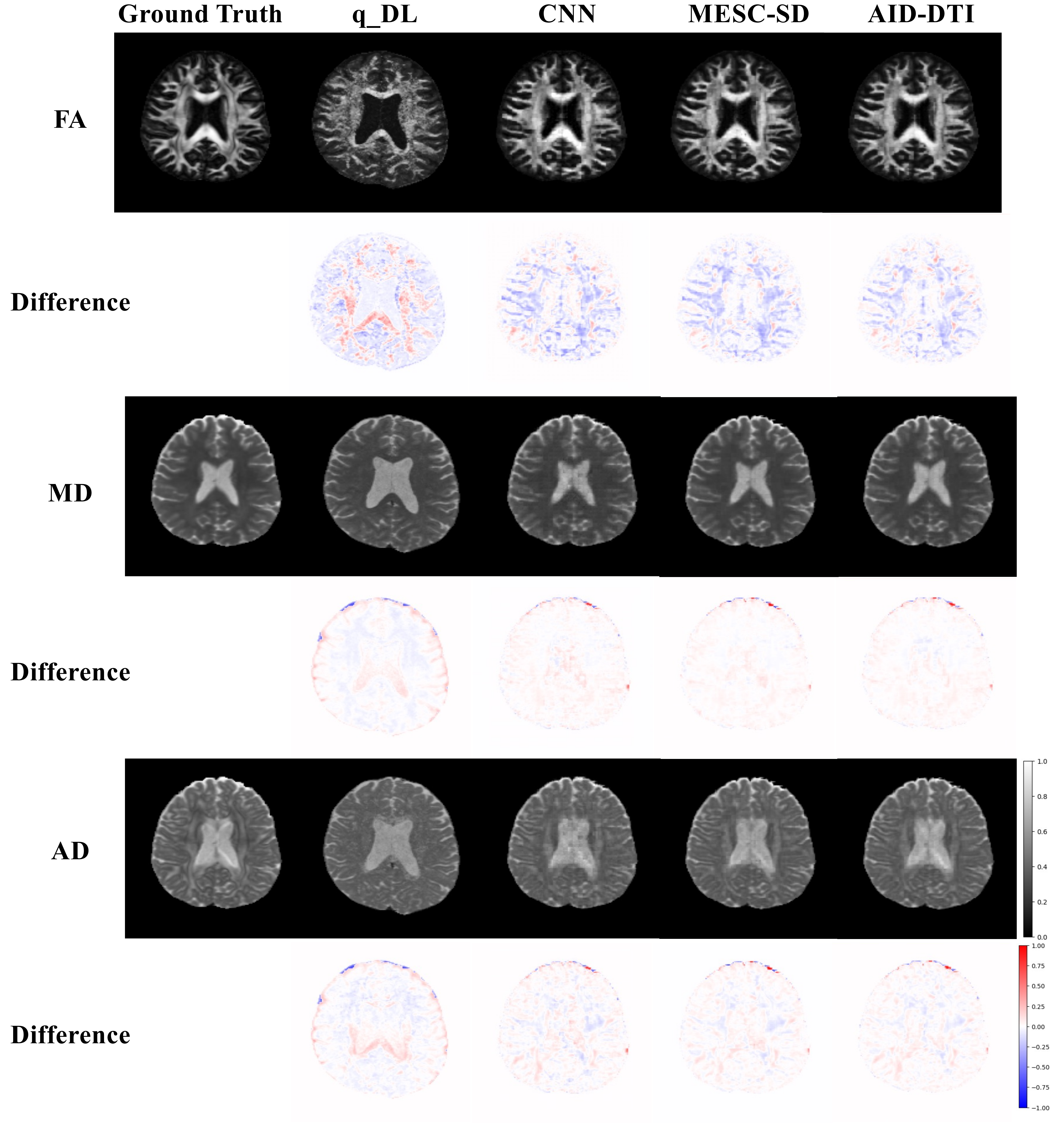}}
  \caption{The ground truth, estimated DTI parameters FA, AD, and MD, and corresponding residual maps based on q-DL, CNN, MESC-SD (baseline), and AID-DTI (ours) in a test subject with 6 diffusion directions at b-values of 1000$s/{mm}^2$.}
\label{fig3}
\end{figure}

In this section, MSE, PSNR and SSIM were used to quantify the similarity compared to the ground truth.
We evaluate the performance of AID-DTI through a comparative analysis between the proposed regularization and its non-regularized version, as well as other state-of-the-art methods. The experimental results are summarized in Table \ref{tab1}, where it can be observed that AID-DTI surpasses the non-regularized methods (MESC-SD \cite{ye2020improved}), q-DL \cite{golkov2016q}, and CNN \cite{gibbons2019simultaneous} by a large margin.
Furthermore, the impact of different learning rates on the effectiveness of each method is investigated. The choice of learning rate is widely recognized as a crucial factor in determining the efficiency and performance of a network. The results presented in Table 1 consistently validate the superiority of AID-DTI across all considered learning rates. 
Illustrative visualization results are shown in Figure \ref{fig3}, which provide further evidence of the improved accuracy and finer details achieved by our proposed method in estimating DTI metrics.

\begin{table}[htp]
\begin{center}
\caption{Quantitative comparison of different methods across various learning rates.}
\label{tab1}
\resizebox{0.5\textwidth}{!}{
\begin{tabular}{c|c c c|c c c|c c c}
\hline
\multirow{2}{*}{Methods}  & \multicolumn{3}{c|}{$lr=0.01$} & \multicolumn{3}{c|}{$lr=0.001$} & \multicolumn{3}{c}{$lr=0.0001$}\\
    \cline{2-10}
    & MSE $\times 10^{-2}$ & PSNR & SSIM & MSE $\times 10^{-2}$ & PSNR & SSIM & MSE $\times 10^{-1}$ & PSNR & SSIM\\
\hline
q-DL\cite{golkov2016q} & 0.2518 & 25.9900 & 0.9472 & 0.2495 & 26.0295 & 0.9482 & 0.2490 & 26.0383 & 0.9484\\
CNN \cite{gibbons2019simultaneous} & 0.2236 & 26.4509 & 0.9513 & 0.1840  & 28.0010 & 0.9582 & 0.2082 & 27.6015 & 0.9546\\
MESC-SD\cite{ye2020improved} & 0.1693 & 27.7138 & 0.9556 & 0.1258 & 29.0016 & 0.9671 & 0.1401 & 28.5365 &0.9624 \\
Ours &\textbf{0.1396} &\textbf{28.5812} &\textbf{0.9626}  &\textbf{0.1214} &\textbf{29.1592} &\textbf{0.9673} &\textbf{0.1281} & \textbf{28.9253} &\textbf{0.9658} \\
\hline
\end{tabular}}
\end{center}
\end{table}

\section{Conclusion and Outlook}
\label{sec:conclusion}

In this study, we develop a novel model-driven deep learning approach AID-DTI for reducing the q-space sampling requirement of DTI to six DWI volumes along uniformly sampled diffusion-encoding directions. Our method maps one $b=0$ image and six DWI volumes to high-quality DTI metrics employing an SVD-based regularization and introduces an adaptive algorithm for automatically updating regularization parameters. The proposed method exhibits simplicity, interpretability, and flexibility and has a high potential to become a practical tool in a wide range of clinical and neuroscientific applications. Future efforts will expand the proposed method to other diffusion models and more multi-parametric MR imaging scenarios.

\section{Acknowledgments}
\label{sec:Acknowledgments}

This research was partly supported by the National Natural Science Foundation of China (62222118, U22A2040), Guangdong Provincial Key Laboratory of Artificial Intelligence in Medical Image Analysis and Application \\(2022B1212010011), Shenzhen Science and Technology Program (RCYX20210706092104034, JCYJ20220531100213029), and Key Laboratory for Magnetic Resonance and Multimodality Imaging of Guangdong Province (2023B1212060052).

\bibliographystyle{IEEEbib}
\bibliography{refs}

\end{document}